\documentclass{article}
\usepackage{spconf,graphicx}
\usepackage{amsmath,mathtools,enumitem, amssymb}
\usepackage{algorithmicx}
\usepackage{algorithm}
\usepackage{algpseudocode}
\usepackage{subfig,caption}
\usepackage{adjustbox}
\usepackage{csquotes}

\algnewcommand{\Initialize}{\State \textbf{Initialize:}}
\def\NoNumber#1{{\def\alglinenumber##1{}\State #1}\addtocounter{ALG@line}{-1}}

\newcommand\Tstrut{\rule{0pt}{2ex}}         
\newcommand\Bstrut{\rule[-0.9ex]{0pt}{0pt}}   
\newcommand{\etal}{\textit{et al}.}
\newcommand{\ie}{\textit{i}.\textit{e}.,\,}

\DeclareMathOperator*{\argmin}{argmin \>}
\DeclareMathOperator*{\argmax}{argmax \>}
\DeclarePairedDelimiter\norm{\lVert}{\rVert}

\newcommand{\fref}[1]{Fig.~\ref{#1}}
\newcommand{\eref}[1]{(\ref{#1})}
\newcommand{\aref}[1]{Algorithm~\ref{#1}}
\newcommand{\tref}[1]{Table~\ref{#1}}

\title{FACE RECOGNITION USING MULTI-MODAL LOW-RANK DICTIONARY LEARNING}

\name{Homa Foroughi\textsuperscript{*}\thanks{\textsuperscript{*}These authors contributed equally to this work}, Moein Shakeri\textsuperscript{*}\footnotemark[1], Nilanjan Ray, Hong Zhang}
\address{Department of Computing Science, University of Alberta, Edmonton, Canada}

\begin{document}
\setlength{\abovedisplayskip}{2pt}
\setlength{\belowdisplayskip}{2pt}
\setlength{\abovedisplayshortskip}{0pt}
\setlength{\belowdisplayshortskip}{0pt} 
\maketitle
\begin{abstract}
Face recognition has been widely studied due to its importance in different applications; however, most of the proposed methods fail when face images are occluded or captured under illumination and pose variations. Recently several low-rank dictionary learning methods have been proposed and achieved promising results for noisy observations. While these methods are mostly developed for single-modality scenarios, recent studies demonstrated the advantages of feature fusion from multiple inputs. We propose a multi-modal structured low-rank dictionary learning method for robust face recognition, using raw pixels of face images and their illumination invariant representation. The proposed method learns robust and discriminative representations from contaminated face images, even if there are few training samples with large intra-class variations. Extensive experiments on different datasets validate the superior performance and robustness of our method to severe illumination variations and occlusion.
\end{abstract}
\section{INTRODUCTION}
\label{sec:intro}
The last decade has witnessed a tremendous progress in face recognition technologies, and great recognition performance has been reported by different methods under some ideal conditions, but most of these methods are not robust to outliers, occlusions, severe illumination and pose variations. In recent years, dictionary learning (DL) algorithms have been successfully applied to different vision tasks including face recognition. DL is a feature learning technique in which, an input signal is represented with a sparse linear combination of dictionary atoms. To alleviate the effects of aforementioned variations, low-rank (LR) matrix recovery has been integrated into the DL framework, and is shown to achieve promising results when corruption existed. LR matrix recovery~\cite{RPCA} was oroginally proposed to recover a LR matrix from corrupted observations, and have succesfuly been applied to applications like background modeling~\cite{BS-LR} and image classification~\cite{Homa-Arxiv-JP}. Li \etal~\cite{D2L2R2} developed a discriminative DL method by combination of the Fisher discrimination and the LR constraint on sub-dictionaries. Zhang \etal~\cite{Structured-LR-DL} presented a structured, sparse and LR representation for image classification by adding a regularization term to the DL objective function. Recently, Foroughi \etal~\cite{Homa-Arxiv-JP} proposed a joint projection and LR-DL method using dual graph constraints for classification of small datasets, which include considerable amount of variations.

In parallel developments, it is well established that information fusion using multiple sources can generally improve the recognition performance, since it provides a framework to combine information from different perspectives that is more tolerant to the errors of individual sources~\cite{Fusion}. To benefit from information fusion, some methods have also successfully incorporated DL technique into the feature learning framework. Monaci \etal~\cite{Multi3} proposed a multi-modal DL algorithm to extract typical templates, which represents synchronous transient structures between multi-modal features.~\cite{Multi4} proposed an uncorrelated multi-view discrimination DL method based on the Fisher discrimination, that jointly learns multiple uncorrelated discriminative dictionaries from different views. Nevertheless, the only work that integrated LR into multi-modal DL was presented by Wu \etal~\cite{Multi5} through constructing class-specific sub-dictionaries for each modality, and utilizing LR and incoherence constraints on each view. 

To construct different modalities, most of the existing methods either exploit multi-view angles~\cite{Multi5} or extract different local features~\cite{Multi5} or weak biometrics~\cite{Multi-task} from pre-defined regions of face images. These methods suffer from two main disadvantages that burden extra overhead on the system. Firstly, they demand either several cameras or manual region definition and hand-crafted feature extraction and secondly, they are not applicable to millions of available face data that have already been captured under single view. By exploiting more meaningful modalities, we address these challenges and even increase the recognition rate further. Recently, Shakeri \etal~\cite{Illum-Invar} presented an illumination invariant representation of an image for outdoor place recognition. To create this representation, they use a Wiener filter derived from the power law spectrum assumption of natural images that is robust against illumination variations. Since the obtained representation may lose the chromaticity of the image, a shadow removal method based on entropy minimization is utilized. This representation showed superior performance for outdoor place recognition in various illumination and shadow variations. Inspired by this success, we design a framework for multi-modal fusion with the following contributions:
\begin{itemize} [leftmargin=0.1in]   
\setlength\itemsep{0em}
\item We design a multi-modal LR-DL method, where in each modality a discriminative and reconstructive dictionary, and a structured sparse and LR representation are learned from face images, and the collaboration between modalities is encouraged by incorporating an ideal representation term. We provide a new classification schema, which utilizes the reconstruction by LR and sparse noise components.
\item By adopting illumination invariant representation of images as one of the modalities, the model learns robust and discriminative representations from noisy images, even when the kind of variation is different in the training and test sets. The proposed method achieves superior performance for small datasets that have large intra-class variation.
\end{itemize}
\section{THE PROPOSED MM-SLDL METHOD}
\label{sec:proposed}
We propose a \textit{M}ulti-\textit{M}odal \textit{S}trcutured \textit{L}ow-rank \textit{D}ictionary \textit{L}earning method (MM-SLDL) for face recognition, in which we use two modalities. While the first modality is constructed by the raw pixels of face images, the second is formed by illumination invariant images~\cite{Illum-Invar}. Denote $X_K \> (K=1,2)$ the training data from the $K^{th}$ modality including $C$ classe, as $X_K=\{ X_K^1, X_K^2, \dots , X_K^C \}$, where $X_K^i$ corresponds to class $i$ in the $K^{th}$ modality. In each modality, we use a supervised learning method to learn a discriminative and reconstructive dictionary $D_K$, and a structural sparse and LR image representation $Z_K$. LR matrix recovery helps to decompose the corrupted matrix $X_K$ into a LR component $DZ$ and a sparse noise component $E$, \ie $X_K = D_K \, Z_K + E_K$. With respect to dictionary $D_K$, the optimal representation matrix $Z_K$ for $X_K$ should be block-diagonal~\cite{LRR}, \ie $Z^{\ast \, ii}_K = Z_K^i$.
In each modality, the dictionary $D_K$ contain $C$ sub-dictionaries as $D_K=\{ D_K^1, D_K^2, \dots , D_K^C \}$, where $D_K^i$ corresponds to the $i^{th}$ class. Let $Z_K^i=\{ Z_K^{i,1}, Z_K^{i,2}, \dots , Z_K^{i,C} \}$ be the representation of $X_K^i$ with respect to $D_K$, then $Z_K^{i,j}$ denotes coefficients for $D_K^j$. To learn robust representations from images, $D_K$ should have discriminative and reconstructive power. Firstly, $D_K^i$ should well represent the samples in class $i$, and ideally be exclusive to each subject $i$. Secondly, every class $i$ needs to be well represented by its sub-dictionary, such that $X_K^i = D_K^i \, Z_K^{i,i} + E_K^i$, and finally $Z_K^{i,j}$, the coefficients for $D_K^j \, (i \neq j)$, are nearly all zero. So, the objective function of MM-SLDL is defined as:
\begin{gather}
\hspace{-10pt}
\min_{D_K,Z_K,E_K} \> \sum\limits_{K=1}^2 ( \norm{Z_K}_* +  \beta \norm{Z_K}_1 + \lambda \norm{E_K}_1) 
\label{eq2} \notag \\ 
 + \alpha \norm{Z_1 Z_2^T-Q}_F^2 \quad s.t. \quad X_K=D_K \, Z_K + E_K \quad K=1,2
\end{gather}
The main objective function simultaneously trains two dictionaries and representations under the joint ideal regularization prior. $Q$ is an ideal representation built from training data in block-diagonal form, and defined as $Q=\{ q_1, q_2, \dots , q_{\mathcal{C}} \} \in R^{\mathcal{C} \times \mathcal{C}}$. Here $\mathcal{C}$ is the size of dictionary, and $q_i$ is the code for sample $x_K^i$ in the form of $[0 \dots p_i, p_i, p_i, \dots]^t \in R^{\mathcal{C}}$, where $p_i$ is the number of training samples in class $i$. This means that if $x_K^i$ belongs to class $\textit{L}$, then the coefficients in $q_i$ for $D_K^{\textit{L}}$ are all $p_i$s, while the others are all $0$s. We add the regularization term $\norm{Z_1 Z_2^T-Q}_F^2$ for two reasons: first, to include the structure information into the dictionary learning process and second, to enforce collaboration between two modalities. It encourages the training images of the same class to have the same representation $Z_K$ in different modalities, despite of intra-class variations.

\textbf{Classification Scheme:} After dictionaries $D_K$ are learned, the LR sparse representations $Z_K$ of training data $X_K$ and $Z_K^{ts}$ of test data $X^{ts}$ are calculated by solving~\eref{eq2} separately using~\aref{alg:MM-SLDL} with $\alpha = 0$. The representation $Z_{K,i}^{ts}$ of the $i^{th}$ test sample is the $i^{th}$ column of $Z_K^{ts}$. Using the multivariate ridge regression model~\cite{Ridge}, we obtain a linear classifier $\hat{W_K}$ as:
\begin{gather}
\hat{W_K} = \argmin_{W_K} \norm{H-W_K Z_K}_2^2 + \lambda \norm{W_K}_2^2 
\label{eq3-0}
\end{gather}
where $H$ is the class label matrix of $X_K$. This yields $\hat{W} = H Z_K^T (Z_K Z_K^T + \lambda I)^{-1}$. The estimated label of the $K^{th}$ modality is obtained as:
\begin{gather}
c_K = \argmax_{c_K} \big( s=(\hat{W_K}+Q)  Z_{K,i}^{ts} \big)
\label{eq3-1}
\end{gather}
where $s$ is the class label vector. We then use LR matrix recovery to obtain LR and sparse noise components of potential classes $c_1,c_2$, and then compute the reconstruction error of the given query sample $X_i^{ts}$ in both modalities:
\begin{gather}
\norm{L(c_K) - \big( X_{K,i}^{ts} - \bar{S}(c_K) \big) }_F^2
\label{eq3-2}
\end{gather}
where $L(c_K)$ is the LR component of class $c_K$ in the $K^{th}$ modality, and $\bar{S}(c_K)$ is the average sparse noise of that class. Since the data range is different in two modalities, we use a normalization step, and the winner class is the one that minimizes the ratio of~\eref{eq3-2} between two modalities.
\section{OPTIMIZATION OF MM-SLDL}
\label{sec:opt}
In each iteration we update the variables of the $K^{th}$ modality, while fixing the variables of other modality, and for the $K^{th}$ modality, the variables are updated alternatively. To solve optimization problem~\eref{eq2}, we first introduce an auxiliary variable $W_K$ to make it separable:
\begin{gather}
\hspace{-10pt}
\min_{D_K,Z_K,E_K} \> \sum\limits_{K=1}^2 (\norm{Z_K}_* +  \beta \norm{W_K}_1 + \lambda \norm{E_K}_1) 
\label{eq4} \\ \notag 
+ \alpha \norm{Z_1 Z_2^T-Q}_F^2  \quad s.t. \quad X_K=D_K \, Z_K + E_K \, , \, W_K = Z_K
\end{gather}
The augmented Lagrangian function $L$ of~\eref{eq4} is defined as:
\begin{gather}
L= \norm{Z_K}_* +  \beta \norm{W_K}_1 + \lambda \norm{E_K}_1 + \alpha \norm{Z_1 Z_2^T-Q}_F^2  \nonumber
\label{eq5}
\end{gather}
\begin{gather}
\hspace{-8pt} + <Y_K,X_K-D_K \, Z_K - E_K> + <M_K,Z_K-W_K> 
\notag \\  
+ \, \frac{\mu}{2} ( \norm{X_K-D_K \, Z_K - E_K}_F^2 + \norm{Z_K-W_K}_F^2 ) \quad K=1,2
\end{gather}
where $<A,B> = \textit{tr}(A^T B)$, $Y_K$ and $M_K$ are Lagrange multipliers and $\mu$ is a balance parameter. The optimization problem~\eref{eq5} can be divided into two sub-problems as follows:
\begin{itemize}[leftmargin=0.1in]
\setlength\itemsep{0em}
\item \textbf{Updating Coding Coefficient $Z_K$:} With $D_K$ fixed, we use the linearized alternating direction method with adaptive penalty (LADMAP)~\cite{LADMAP} to solve for $Z_K$ and $E_K$. The augmented Lagrangian function~\eref{eq5} would reduce to:
\begin{gather}
\norm{Z_K}_* +  \beta \norm{W_K}_1 + \lambda \norm{E_K}_1 + \alpha \norm{Z_1 Z_2^T-Q}_F^2 
\label{eq6} \notag \\  
+ \frac{\mu}{2} ( \norm{X_K-D_K \, Z_K - E_K + \frac{Y_K}{\mu}}_F^2 
\notag \\ 
+ \norm{Z_K-W_K+  \frac{M_K}{\mu}}_F^2 ) - \frac{1}{2\mu} (\norm{Y_K}_F^2 + \norm{M_K}_F^2)
\end{gather}
The function~\eref{eq6} should be minimized by alternative updating variables $Z_K,W_K,E_K$ as follows:
\begin{gather}
Z_K^{j+1} = \argmin_{Z_K} \frac{1}{\eta \mu} \norm{Z_K}_* +  \frac{1}{2} \norm{Z_K - Z_K^j + \mu [-D_K^T (X_K 
\label{eq7} \notag \\ 
\hspace{-3pt} - D_K \, Z_K^j - E_K^j +\frac{Y_K^j}{\mu}) + \frac{2 \alpha (Z_K Z_l^T -Q) Z_l }{\eta \mu} 
\notag \\  
+ \mu (Z_K - W_K^j +\frac{M_K^j}{\mu}) ] / \eta \mu }_F^2 \quad \text{where} \quad l \neq K
\end{gather}
where $\eta = \norm{D_K}_2^2$, and we notice that $Q=Q^T$.
\begin{gather}
W_K^{j+1} = \argmin_{W_K} \frac{\beta}{\mu} \norm{W_K}_1 +  \frac{1}{2} \norm{W_K - Z_K^{j+1} - \frac{M_K^j}{\mu} }_F^2
\label{eq8} 
\end{gather}
\begin{gather}
E_K^{j+1} = \argmin_{E_K} \frac{\lambda}{\mu} \norm{E_K}_1 +  \frac{1}{2} \norm{E_K -  (\frac{Y_K^j}{\mu}  + X_K 
\label{eq9} \notag \\  
-D_K Z_K^{j+1}) }_F^2
\end{gather}
\item \textbf{Updating Dictionary $D_K$:} When $Z_K,W_K,E_K$ are fixed, we would be able to update $D_K$. The Lagrangian function~\eref{eq5} is further reduced to:
\begin{gather}
\frac{\mu}{2} (\norm{X_K-D_K \, Z_K - E_K + \frac{Y_K}{\mu}}_F^2 + \norm{Z_K-W_K}_F^2)
\label{eq10} \notag \\  
 + C(Z_K,W_K,E_K,Q) 
\end{gather}
where $C(Z_K,W_K,E_K,Q) $ is fixed. Equation~\eref{eq10} is in the quadratic form and $D_K$ can be solved directly as follows:
\begin{gather}
D_K^{j+1} = \gamma D_K^j + (1-\gamma) D_K^{update}
\label{eq11} 
\end{gather}
where 
$D_K^{update} = \frac{1}{\mu} (Y_K + \mu (X_K - E_K)) Z_K^T (Z_K Z_K^T)^{-1} $. We initialize the dictionary using KSVD method on training samples of each class and combining all the classes.
\begin{algorithm}[!ht]
\footnotesize
\caption{MM-SLDL Method in the $K^{th}$ Modality}
\label{alg:MM-SLDL}
\begin{algorithmic}[1]
\Require Data $X_K$, Parameters $\lambda, \beta, \alpha, \gamma$
\Ensure $D_K, Z_K$
\Initialize $\, D_K^0 ; \, Z_K^0 = W_K^0 = E_K^0 = Y_K^0 = M_K^0 = 0 ; \, \mu=10^{-6} ; \, max_{\mu} = 10^{30}; \, \epsilon_s = 10^{-8}; \, \rho =1.1 ; \, \epsilon_d = 10^{-5}$
\While {not converged}
    \State Fix other variables and update $Z_K$ by Equation~\eref{eq7} 
    \State Fix other variables and update $W_K$ by Equation~\eref{eq8} 
    \State Fix other variables and update $E_K$ by Equation~\eref{eq9} 
    \State Update $Y_K,M_K$ as: 
    \NoNumber {$\hspace{45pt} Y_K = Y_K + \mu (X_K-D_K \, Z_K - E_K)$}
    \NoNumber {$\hspace{45pt} M_K = M_K + \mu (Z_K - W_K)$} 
    \State Update $\mu$ as: $\mu = min(\rho \mu, max_{\mu})$    
    \State Check stopping conditions as: 
    \NoNumber {$\hspace{-15pt} \norm{X_K-D_K \, Z_K - E_K}_{\infty} < \epsilon_s \quad \text{and} \quad \norm{Z_K - W_K}_{\infty} < \epsilon_s $}
\EndWhile
\While {not converged}
    \State Fix other variables and update $D_K$ by Equation~\eref{eq11} 
    \NoNumber {$\hspace{35pt} \norm{D_K^{j+1} - D_K^j}_{\infty} < \epsilon_d $}
\EndWhile
\end{algorithmic}
\end{algorithm}
\end{itemize}
\section{RESULTS AND DISCUSSION}
\label{sec:exp}
The performance of MM-SLDL method is evaluated on three face datasets. We compare our method with three types of methods: (1)Multi-modal LR-DL method MLDL~\cite{Multi5} (2)Multi-modal DL methods including UMD\textsuperscript{2}L~\cite{Multi4} and MSDL~\cite{Multi1} (3)Single-modality LR-DL methods such as JP-LRDL~\cite{Homa-Arxiv-JP}, D\textsuperscript{2}L\textsuperscript{2}R\textsuperscript{2}~\cite{D2L2R2} and SLRDL~\cite{Structured-LR-DL}. For constructing the training set, we select images randomly and the selection is repeated $10$ times and we report the average recognition rates for all methods. We set the number of dictionary atoms of each class as training size, and choose the tuning parameters of all methods by 5-fold cross validation.
\begin{table}[b]
\caption{Recognition rates (\%) on AR dataset}
\fontsize{8.5}{8}\selectfont
\vspace{-0.5em}
\label{table:AR}
\centering
\begin{tabular}{|l||c|c|c|c|}
\hline
Method & Sunglasses & Scarf & Mixed & Misc.\Tstrut\Bstrut\\
\hline \hline 
MLDL~\cite{Multi5}                    & 90.51 & 91.51 & 91.32 & 76.33 \Tstrut\Bstrut\\
UMD\textsuperscript{2}L~\cite{Multi4} & 88.26 & 87.40 & 88.30 & 71.30 \Tstrut\Bstrut\\
MSDL~\cite{Multi1}                    & 83.20 & 80.65 & 79.50 & 68.44 \Tstrut\Bstrut\\
D\textsuperscript{2}L\textsuperscript{2}R\textsuperscript{2}~\cite{D2L2R2} & 92.20 & 90.40 & 91.30 & 75.30 \Tstrut\Bstrut\\
SLRDL~\cite{Structured-LR-DL}         & 87.35 & 83.40  & 82.47 & 72.30  \Tstrut\Bstrut\\
JP-LRDL~\cite{Homa-Arxiv-JP}          & 93.20 & 93.00  & 93.30 & 78.23  \Tstrut\Bstrut\\
AlexNet~\cite{Alex-Net}               & 30.33 & 30.12  & 30.17 & 25.55  \Tstrut\Bstrut\\
VGG-Face~\cite{VGG-Face}              & 85.90 & 85.01  & 87.30 & 79.83  \Tstrut\Bstrut\\
\textbf{MM-SLDL}                      & \textbf{96.70} & \textbf{96.41} & \textbf{96.30} & \textbf{85.30}\Tstrut\Bstrut\\
\hline
\end{tabular}
\vspace{-1.5em}
\end{table}
Convolutional Neural Networks (CNNs) have significantly improved the face recognition rates, and the most important ingredient for the success of such methods is the availability of large quantities of training data; however, transfer learning is a powerful tool to train small target datasets.~\cite{Transferable} revealed when the target dataset is small and similar to original dataset, it is better to treat CNN as fixed feature extractor and train a linear classifier on the CNN features. We compare MM-SLDL with two deep methods (1)Deep features generated by VGG-Face descriptor~\cite{VGG-Face}, that is based on a $16$-layer CNN trained on $2.6$M images, followded by a nearest neighbor classifier. (2)We use $8$-layer AlexNet~\cite{Alex-Net} trained on $1.2$M images of the ImageNet dataset, and fine-tune it on the target data.

\textbf{AR Dataset}~\cite{AR} includes over $4,000$ face images from $126$ individuals, $26$ images for each person in two sessions. Among the images of each session, $3$ are obscured by scarves, $3$ by sunglasses, and the remaining faces are of different facial expressions or illumination variations, which we refer to as unobscured images. Following~\cite{Structured-LR-DL}, experiments are conducted under three scenarios. \textbf{Sunglasses:} We select $7$ unobscured images and $1$ with sunglasses from the first session as training samples for each person, and the rest of unobscured and sunglasses images are used for testing. \textbf{Scarf:} We choose $8$ training images ($7$ unobscured and $1$ with scarf) from the first session for training, and $12$ test images including the rest of unobscured and scarf images. \textbf{Mixed:} We select $7$ unobscured, plus $2$ occluded images ($1$ with sunglasses, $1$ by scarf) from the first session for training, and the remaining $17$ images in two sessions for testing. We design a challenging scenario \textbf{Misc.}, in which we select $7$ unobscured, and $1$ scarf images from the first session for training, and utilize the remaining $7$ unobscured and $6$ sunglasses images for testing. Here, the type of noise is different in training and test sets. According to~\tref{table:AR}, MM-SLDL achieves the best performance in all scenarios, and the improvement is significant in \enquote{Misc.} scenario, where all the other methods fail.
\begin{figure}[t]
\centering
\subfloat[Classification]{\includegraphics[width=5.1cm,height=2.1cm]{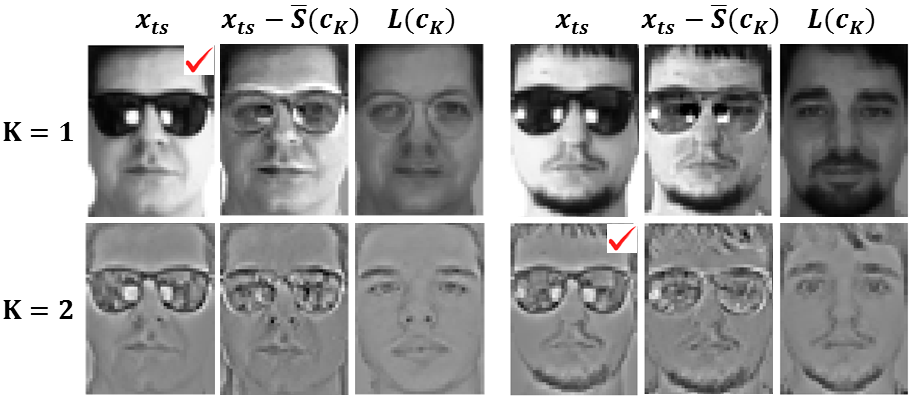} \label{fig:AR-Classification}}
\hspace{1pt}
\subfloat[Decomposition]{\includegraphics[width=3cm,keepaspectratio]{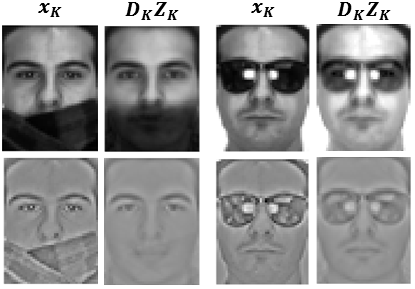} \label{fig:AR-Decomposition}}  
\vspace{-0.5em}
\caption{Image decomposition and classification on AR dataset}
\vspace{-1.5em}
\end{figure} 
\begin{figure}[h]
\centering
\subfloat[]{\adjustbox{raise=0.9pc}{\includegraphics[width=2cm,height=3.3cm]{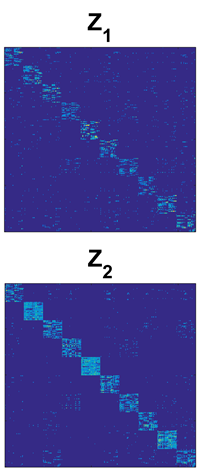} \label{fig:Z-Yale}}} 
\hspace{10pt}
\subfloat[]{\includegraphics[width=4.4cm,keepaspectratio]{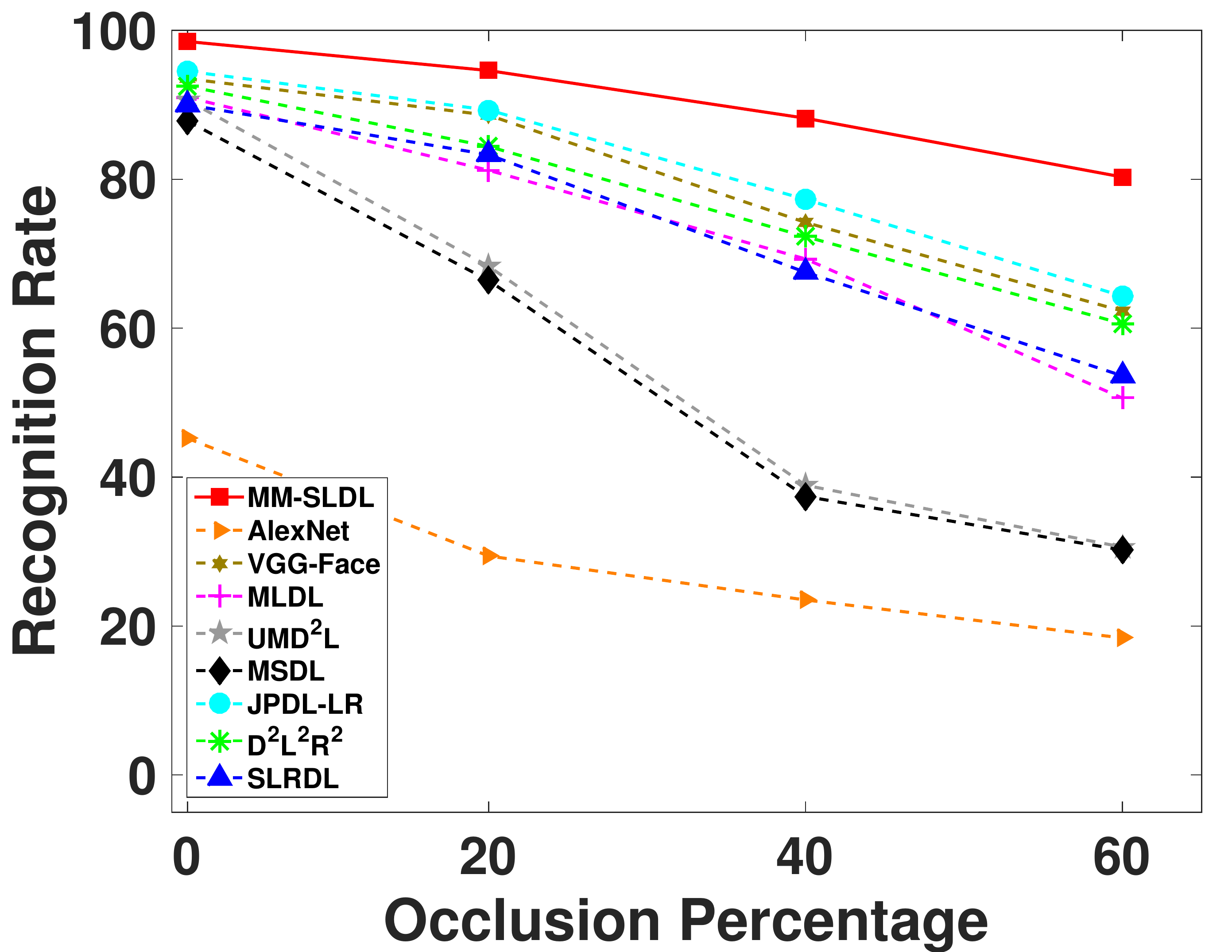} \label{fig:Yale-Results}}  
\vspace{-0.5em}
\caption{(a) Representations for testing samples and (b) Recognition rates (\%) on Extended YaleB dataset}
\vspace{-0.5em}
\end{figure} 
\fref{fig:AR-Decomposition} illustrates examples of image decomposition on AR dataset. The first and second rows show training images and learned LR component $D_K Z_K$ in two modalities. While the first modality keeps more details, the illumination invariant modality better separates occlusions from the original images; hence, a robust representation is learned by their fusion.~\fref{fig:AR-Classification} demonstrates a testing sample $x_{ts}$, and $x_{ts} - \bar{S}(c_K),L(c_K)$ components, which their difference determines the winner, that is illustrated by a red tick mark.

\begin{figure}[t]
\centering
\subfloat[Extended YaleB]{\includegraphics[width=3.5cm,keepaspectratio]{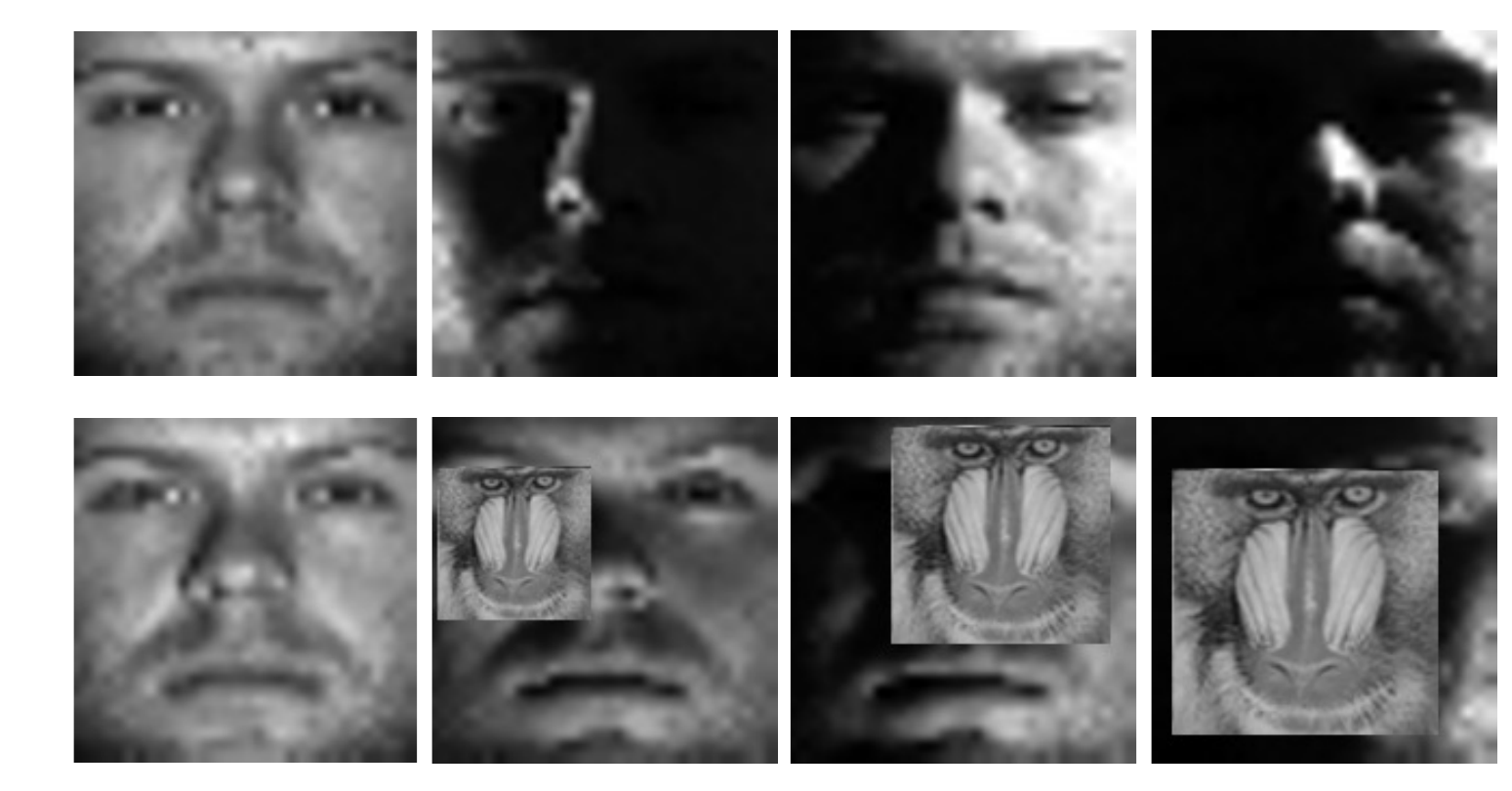} \label{fig:Yale-Samples}}  
\hspace{3pt}
\subfloat[LFWa]{\includegraphics[width=3.5cm,keepaspectratio]{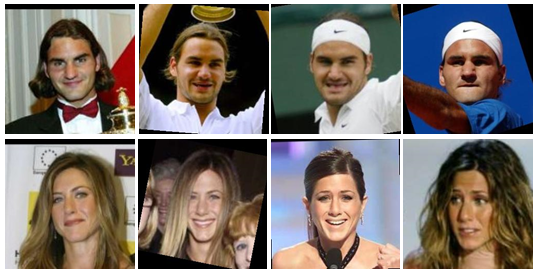} \label{fig:LFW-Samples}}  
\vspace{-0.5em}
\caption{Sample images of Extended YaleB and LFWa datasets}
\vspace{-1.5em}
\end{figure}

\textbf{Extended YaleB Dataset}~\cite{Yale} contains $2,414$ face images of $38$ human subjects captured under different illumination conditions. There are $59 \sim 64$ images for each subject, and we randomly select $20$ of them for training. We simulate various levels of contiguous occlusion from $20\%$ to $60\%$, by replacing a randomly located square block of each train image with an unrelated image, as seen in~\fref{fig:Yale-Samples}. To have a real challenge, test images are not occluded. We visualize the representation $Z$ for two modalities for testing images of the first $10$ classes under $40\%$ occlusion training scenario in~\fref{fig:Z-Yale}. Testing images automatically generate a block diagonal structure, and the second modality learns a better representation here.~\fref{fig:Yale-Results} illustrates the recognition rates of all methods across different occlusion levels, and MM-SLDL outperforms other counterparts, especially for severely occluded images.

\textbf{LFW Dataset}~\cite{LFW} contains $13,233$ unconstrained face images of $5,749$ different individuals, collected from the web with large variations in pose, expression, illumination, clothing, hairstyles, occlusion, etc. We use an aligned version of LFW called LWFa~\cite{LFW-a}, and exploit $143$ subject with no less than $11$ samples per subject to perform the experiment. Some of these images are shown in~\fref{fig:LFW-Samples}. A central $170 \times 140$ region is cropped from each of images and the first $10$ samples per class are selected for training, while the rest is used for testing.~\tref{table:LFW-Results} shows the recognition rates of all compared methods. Although VGG-Face has already been trained on extra $2.6$M images, MM-SLDL achieves competitive results using too much smaller training data, which have large intra-class variation. Also, as expected fine-tuned AlexNet is prone to overfitting because target data is small and very different in content compared to the ImageNet. Finally, to verify the role of illumination invariant modality, we just use $K=2$ in the objective function~\eref{eq2} and change the ideal representation term to $\norm{Z_K-Q}_F^2$. The results are reported under \textit{SLDL-Mod2}, and as observed is not competitive.
\begin{table}[h]
\caption{Recognition rates (\%) on LFWa dataset}
\fontsize{8.5}{8}\selectfont
\vspace{-0.5em}
\label{table:LFW-Results}
\centering
\begin{tabular}{|l|c|l|c|}
\hline
Method & Rec. Rate & Method & Rec. Rate \Tstrut\Bstrut\\
\hline \hline 
MLDL~\cite{Multi5}              & 74.10 &   UMD\textsuperscript{2}L~\cite{Multi4}      & 70.43 \Tstrut\Bstrut\\
MSDL~\cite{Multi1}              & 64.25 &   D\textsuperscript{2}L\textsuperscript{2}R\textsuperscript{2}~\cite{D2L2R2}  & 75.20 \Tstrut\Bstrut\\
SLRDL~\cite{Structured-LR-DL}   & 74.20 &   JP-LRDL~\cite{Homa-Arxiv-JP} & 79.87 \Tstrut\Bstrut\\
AlexNet~\cite{Alex-Net}         & 40.31 &  VGG-Face~\cite{VGG-Face}     & 90.01  \Tstrut\Bstrut\\
\textit{SLDL-Mod2}              & 76.77 &  \textbf{MM-SLDL}   & \textbf{88.04}  \Tstrut\Bstrut\\ 
\hline
\end{tabular}
\vspace{-1.5em}
\end{table}
\section{CONCLUSIONS}
\label{sec:con}
We proposed a face recognition method that learns discriminative dictionaries and structured sparse LR representations from contaminated face image in two modalities. Adopting the illumination invariant representation of images as a modality, additionally empowers the model. Experimental results indicate that MM-SLDL is robust, achieving state-of-art performance in the presence of occlusion, illumination and pose changes, using a few training samples. 
\bibliographystyle{IEEEbib}
\bibliography{Homa_Submission}
\end{document}